\documentclass[letterpaper,twocolumn,fleqn]{article} 

\usepackage{ist}
\usepackage{epsfig}
\usepackage{graphicx}
\usepackage{amsmath}
\usepackage{amssymb}
\usepackage{graphicx}
\usepackage{booktabs}
\usepackage{balance}
\usepackage{multirow}
\usepackage{algorithm2e}
\usepackage{adjustbox}

\usepackage{url}

\title{Improving Video Deepfake Detection: A DCT-Based Approach with Patch-Level Analysis}

\author{Luca Guarnera, Salvatore Manganello, Sebastiano Battiato; Department of Mathematics and Computer Science, University of Catania, Italy}

\date{} 

\begin{document} 

\maketitle 

\thispagestyle{empty} 


\begin{abstract}

A new algorithm for the detection of deepfakes in digital videos is presented. The I-frames were extracted in order to provide faster computation and analysis than approaches described in the literature. To identify the discriminating regions within individual video frames, the entire frame, background, face, eyes, nose, mouth, and face frame were analyzed separately. From the Discrete Cosine Transform (DCT), the $\beta$ components were extracted from the AC coefficients and used as input to standard classifiers. Experimental results show that the eye and mouth regions are those most discriminative and able to determine the nature of the video under analysis.

\end{abstract}

\section{Introduction}

The rapid development of the deepfake technology poses significant challenges to the authenticity and trustworthiness of multimedia content~\cite{battiato2016multimedia}. Deepfakes are synthetic creations that manipulate or generate content using advanced generative models, making it increasingly difficult to differentiate them from genuine recordings. This has raised concerns regarding the potential misuse of deepfakes for malicious purposes, such as disinformation campaigns, identity theft, or defamation. The rapid advancement of deep learning techniques, coupled with the availability of vast amounts of data, has led to increasingly sophisticated deepfake algorithms.

The Deepfake phenomenon first emerged in 2017 on the website ``Reddit", when an anonymous user named ``deepfakes" uploaded a pornographic video that superimposed a celebrity's face onto another person's body. This event sparked the interest of other users, who began creating similar videos by replacing the faces of different celebrities. Research indicates that approximately 96\% of Deepfake videos are pornographic in nature, while the remaining 4\% cover various other genres.
Deepfake technology primarily focuses on manipulating faces, as facial recognition plays a crucial role in identifying individuals. As a result, four main face manipulation modalities are employed: generation of entire synthetic faces, facial attribute manipulation, identity swapping, and expression swapping.

\begin{figure}[t!]
    \centering
\includegraphics[width=1\linewidth]{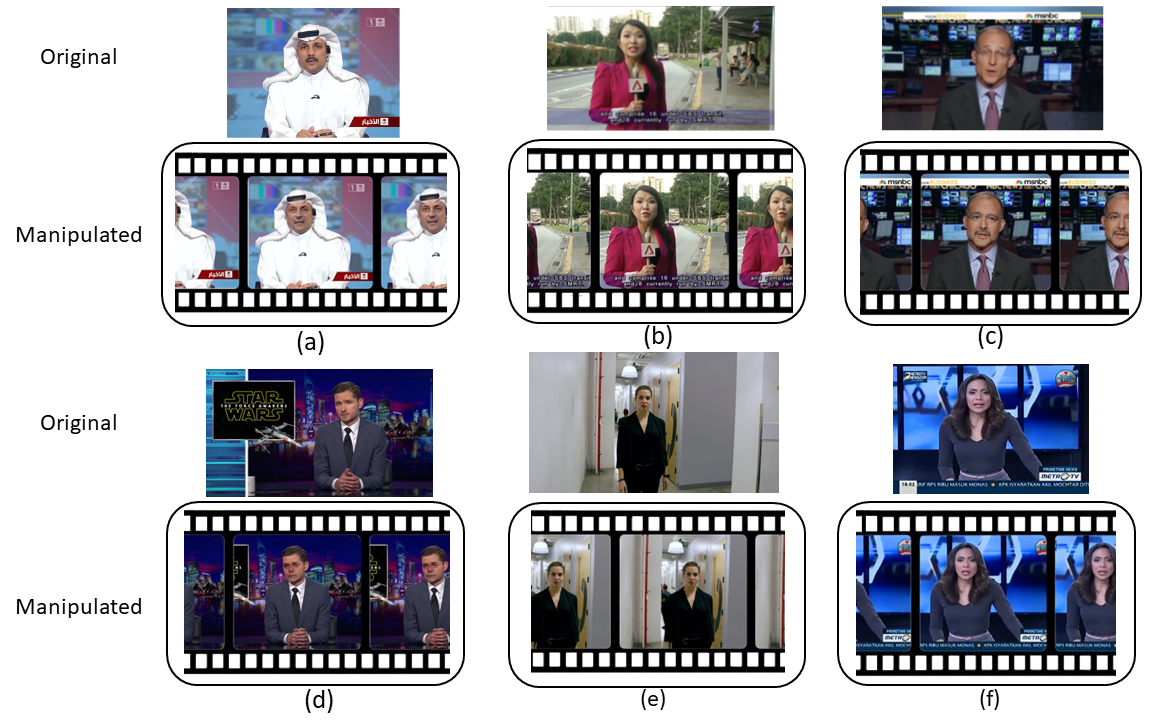}
    \caption{Videos in the Faceforensics++ dataset manipulated with respect to the techniques of (a) Faceswap, (b) Face2Face, (c) Face Shifter, (d) DeepFakes, (e) DeepFake Detection, (f) Neural textures.}
    \label{fig:FFvideo}
\end{figure}

Examples of deepfake videos have demonstrated the potential for creating realistic yet fabricated scenarios. For instance, politicians have been depicted giving speeches they never delivered, celebrities have been inserted into explicit scenes, and individuals' faces have been swapped onto others in various contexts. Examples include the ``Synthesizing Obama" project~\footnote{\url{https://www.youtube.com/watch?v=cQ54GDm1eL0}, last accessed 10/03/2023.}, which is a fake video showing former US President Barack Obama and American director Jordan Peele saying words and phrases they have never said using a technique called lip-syncing. The face of Nicolas Cage has also been put in films where the same actor did not act, such as ``Matrix" or ``Fight Club", as well as in music videos such as ``Never Gonna Give You Up"~\footnote{\url{https://www.youtube.com/watch?v=4soZciRrZRI}, last accessed 13/03/2023.}. These examples highlight the potential consequences of deepfake technology, including its potential to deceive, manipulate public opinion, and compromise personal privacy. 

To address the risks associated with deepfakes, the development of reliable and efficient deepfake detection methods has become necessary. Researchers have explored various approaches, including deep learning models~\cite{xuan2019generalization,hsu2020deep,gandhi2020adversarial,li2020face,wang2020cnn}, statistical analysis~\cite{guarnera2020deepfake,guarnera2020fighting,mccloskey2019detecting,giudice2021fighting}, and forensic~\cite{guarnera2022deepfake} techniques, to detect the presence of deepfakes in multimedia content. However, the constant evolution of deepfake algorithms necessitates ongoing research and innovation in this field.
In this study, we propose a novel deepfake detection algorithm that exploits the Discrete Cosine Transform (DCT) to extract discriminative features from video frames. By analyzing the $\beta$ components derived from the AC coefficients, we aim to identify the most informative frequencies for differentiating between real and deepfake videos. Our methodology focuses on both accuracy and computational efficiency, with the goal of enabling real-time deepfake detection in practical scenarios.
Moreover, to enhance the efficacy of our algorithm, we investigated the discriminative patches within individual frames. By analyzing I-frames, we aimed to identify specific regions, such as the eyes and mouth, that exhibit distinctive characteristics when comparing real and deepfake videos. This patch-level analysis further enhances the explainability and reliability of our deepfake detection system.

To evaluate the performance of our algorithm, we conducted extensive experiments on widely used deepfake datasets, including Faceforensics++ and Celeb-DF (v2). Our results demonstrate the effectiveness of the proposed approach in accurately classifying real and deepfake videos.
By presenting our methodology, experimental results, and analysis, this research contributes to the ongoing efforts in combating deepfakes. Our study aims to advance the state-of-the-art in deepfake detection, with a focus on speed, interpretability, and accuracy. The insights gained from this work can help mitigate the potential harmful effects of deepfakes and contribute to the development of robust defense mechanisms against their malicious use.


\section{Related Works}
\label{sec:sota}
Deepfake video detection has become a crucial research area due to the increasing prevalence and potential harm associated with manipulated videos. Several methods have been proposed to tackle this challenge. 
An overview on Media forensics with a particular focus on Deepfakes has been proposed in~\cite{verdoliva2020media,guarnera2020preliminary,zhang2022deepfake}.
Various existing methods have been developed for Deepfake detection, aiming to distinguish genuine multimedia content from Deepfakes. Many approaches focus on identifying traces left by generative models during the creation of Deepfakes. Matern et al. \cite{matern2019exploiting} exploit the presence of artifacts, such as differently colored eyes or differently shaped ears, which are often produced by generative models. Their method involves training two classifiers that extract features to describe these anomalies. However, this technique is limited to synthetic faces with open eyes or visible teeth.


Recent studies have shown that artifacts can also be detected using Convolutional Neural Networks (CNNs) \cite{wang2020cnn,guarnera2023level}. Nguyen et al. \cite{nguyen2019multi} propose a multi-task learning approach, training a CNN for both manipulation detection and segmentation of manipulated areas. Nirkin et al. \cite{nirkin2021deepfake} introduce a CNN-based method that analyzes the face and its relationship with the surrounding context, comparing the extracted features to identify any discrepancies. Haliassos et al. \cite{haliassos2021lips} present the LipForensics method, which leverages CNNs to identify irregularities in lip movements commonly found in synthetic videos. They extract lip features using a CNN and compare them with the rest of the face. Zhang et al. \cite{zhang2021detecting} propose the use of a 3-Dimensional Convolutional Neural Network (3DCNN) to capture spatio-temporal information from videos, enabling differentiation between original and Deepfake videos.

Zheng et al. \cite{zheng2021exploring} propose a two-phase method for Deepfake detection. In the first phase, they introduce a novel architecture called the Fully Temporal Convolution Network (FTCN), which reduces the spatial convolutional matrix dimension to 1 while maintaining the temporal convolutional matrix dimension. In the second phase, a Temporal Transformer network is employed to verify long-term temporal coherence. Ge et al. \cite{ge2022deepfake} propose the Latent Pattern Sensing (LPS) model, which captures semantic change features for Deepfake video detection. LPS involves an analyzer to detect faces, an encoder to extract spatial semantic features, an aggregator to obtain spatiotemporal features, and a classifier to distinguish real videos from Deepfake videos. Also, two additional models, namely LPS${ri}$ and LPS${ssd}$, have been introduced.

\section{Dataset}
\label{sec:dataset}

\begin{figure}[t!]
    \centering
    \includegraphics[width=\linewidth]{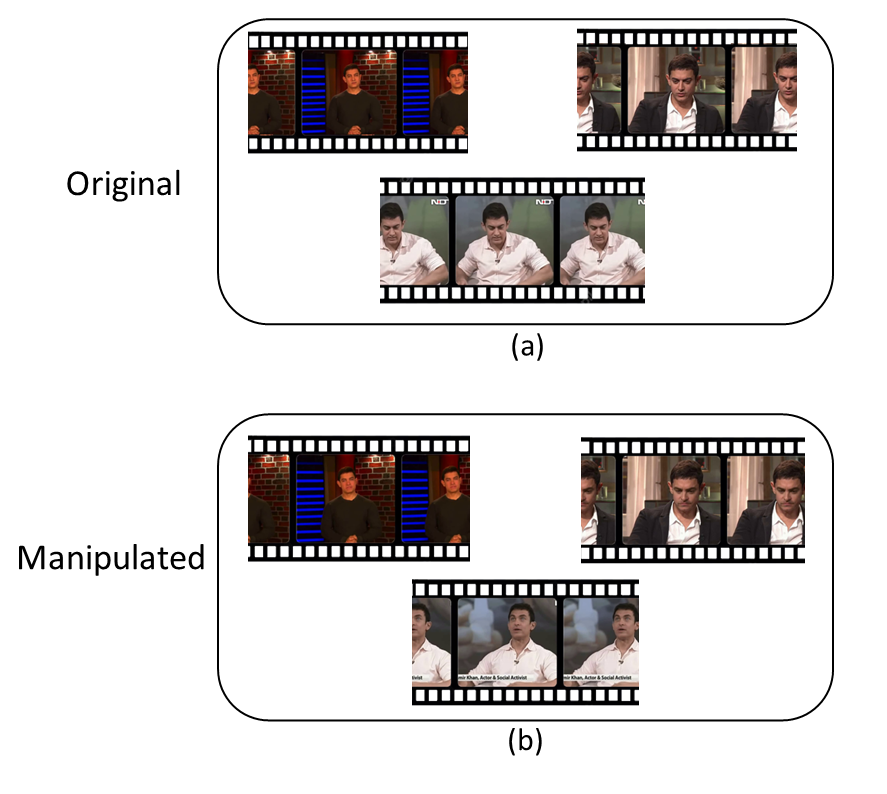}
    \caption{Examples of real (a) and manipulated (b) videos in the Celeb-DF dataset (v2).}
    \label{fig:Celebvideo}
\end{figure}

The experiments were performed using two different types of Deepfake video datasets: \textit{Faceforensics++} \cite{rossler2019faceforensics++} and \textit{Celeb-DF (v2)} \cite{li2019celeb}. Both datasets consist of a combination of real and synthetic videos. The real videos included in both Celeb-DF (v2) and Faceforensics++ were sourced from the internet, primarily from platforms like YouTube, or captured using real actors. The synthetic videos in Faceforensics++ were generated using two types of manipulations: computer graphics-based approaches such as Face2Face, FaceSwap, and learning-based approaches like Deepfakes and NeuralTextures. Figure~\ref{fig:FFvideo} shows some examples of video frames of the FaceForensics++ dataset. 
The Celeb-DF (v2) dataset contains synthetic videos created using publicly available algorithms specifically designed for Deepfake generation. Figure~\ref{fig:Celebvideo} shows some examples. 
Table \ref{tab:dataset} provides an overview of the number of videos included in each dataset.


\begin{table}[t!]
\centering
    \begin{adjustbox}{max width=0.45\textwidth}
\begin{tabular}{cccc}
\hline
\textbf{Dataset}                 & \multicolumn{2}{c}{\textbf{Type}}                                                                    & \textbf{\#Videos} \\ \hline
\multirow{8}{*}{Faceforensics++} & \multirow{6}{*}{Manipulated Sequence} & DeepFakeDetection & 3084              \\ \cline{3-4} 
                                 &                                                                                  & Deepfakes         & 1001              \\ \cline{3-4} 
                                 &                                                                                  & Face2Face         & 1001              \\ \cline{3-4} 
                                 &                                                                                  & NeuralTextures    & 1002              \\ \cline{3-4} 
                                 &                                                                                  & FaceSwap          & 1000              \\ \cline{3-4} 
                                 &                                                                                  & FaceShifter       & 1001              \\ \cline{2-4} 
                                 & \multirow{2}{*}{Original Sequance}     & Actor             & 364               \\ \cline{3-4} 
                                 &                                                                                  & Youtube           & 1003              \\ \hline
\multirow{3}{*}{Celeb-DF (v2)}   & \multirow{2}{*}{Original}                                                        & Celeb-real        & 590               \\ \cline{3-4} 
                                 &                                                                                  & Youtube           & 300               \\ \cline{2-4} 
                                 & Manipulated                                                                      & Celeb-synthesis   & 5639              \\ \hline
\end{tabular}
    \end{adjustbox}
    \caption{Tab.1: Number of videos present in each folder of the utilized dataset.}
    \label{tab:dataset}
\end{table}

\begin{figure*}[t!]
    \centering
    \includegraphics[width=1\linewidth]{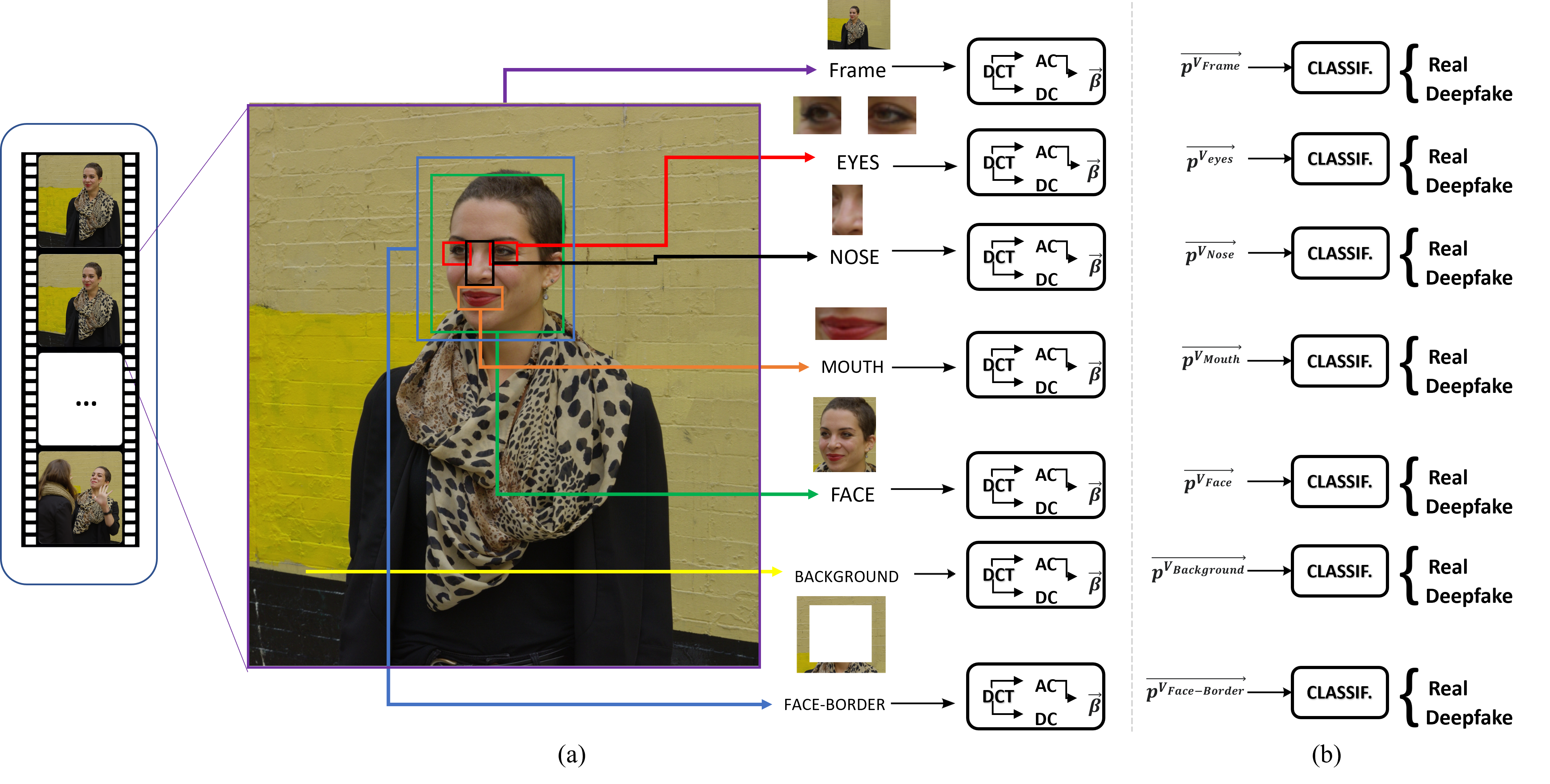}
    \caption{Proposed approach: (a) For each patch in \text{A} of the I-frames of the video $V$, the DCT is calculated and the $\beta$ components of the 63 AC coefficients are extracted. (b) The final feature vectors $p^{V_a}$ of video \textit{V}, are used in the various classifiers to solve the Real Vs Deepfake task and identify the most discriminative regions.}
    \label{fig:p2}
\end{figure*}

In our study, we wanted to conduct a thorough evaluation of deepfake detection methods using a variety of standard classifiers (e.g., k-NN, SVM and others). In the experiments carried out, we trained all classifiers considering both datasets just described. Specifically, that dataset was divided into Training (50\%), Validation (20\%), and Test (30\%), ensuring that the classes of real that of deepfakes were balanced in terms of the number of videos in each category.

Overall, the inclusion of these diverse datasets allows us to evaluate the effectiveness and robustness of our proposed deepfake detection methods across a wide range of scenarios and manipulation techniques.

\section{Proposed Approach}
\label{sec:method}

The proposed pipeline in this paper consists of working on frames of type I (I-frame) of the datasets under analysis. The main objective was to search for anamalies in the Discrete Cosine Transform (DCT) domain by analysis of the $\beta$ parameters extracted from the 63-AC coefficients, which were found to be highly discriminative in distinguishing Deepfake images from real ones~\cite{concas2022tensor,giudice2021fighting}. 
In order to achieve a fast method able to identify the most discriminative area in a video frame, various regions were analyzed separately: 
\begin{center}
    \textit{A = \{entire frame, face, face contour, eyes, nose, mouth, background\}}
\end{center}

Each region of interest in each frame was initially converted to grayscale and then divided into 8x8 blocks. DCT was applied to each block by considering the Formula~\ref{eq:DCT}.

\begin{equation}   
\begin{split}
    F_{u,v}=\frac{2}{N}\biggl[\sum_{x=0}^{N-1}\sum_{y=0}^{N-1}C{(u)}C{(v)}f(x,y) \\
    cos\frac{(2x+1)v\pi}{2 * N}cos\frac{(2y+1)v\pi}{2 * N}\biggl]
    \end{split}
    \label{eq:DCT}
\end{equation}
where $N=8$, $C{(u)}$ and $C{(v)}$ are defined as follows:

\begin{equation}
    C(u) = \bigg \{
\begin{array}{rl}
\frac{1}{\sqrt{2}} & u = 0 \\
1 & u > 0\\
\end{array},
C(v) = \bigg \{
\begin{array}{rl}
\frac{1}{\sqrt{2}} & v = 0 \\
1 & v > 0\\
\end{array} 
\end{equation}

DCT coefficients are generated and ordered in zig-zag order from the top-left element to the bottom-right element. From the DCT, it is possible to analyze the DC and AC coefficients, where the DC coefficient represents the average luminance value and is considered a redundant value. All other coefficients are AC and represent specific frequency bands. The DC coefficient follows a Gaussian distribution while the AC coefficients follow a Laplacian distribution centered at 0, according to the following Formula~\ref{eq:lapla}:
 \begin{equation}
    P(x)= \frac{1}{2\beta}exp\biggl(\frac{-\left| x - \mu \right|}{\beta}\biggl)  
    \label{eq:lapla}
\end{equation}

where $\mu=0$, $\beta=\sigma/\sqrt{2}$ and $\sigma$ corresponds to the standard deviation of the AC coefficient. 

\begin{figure*}[t!]
    \centering
    \includegraphics[width=1\linewidth]{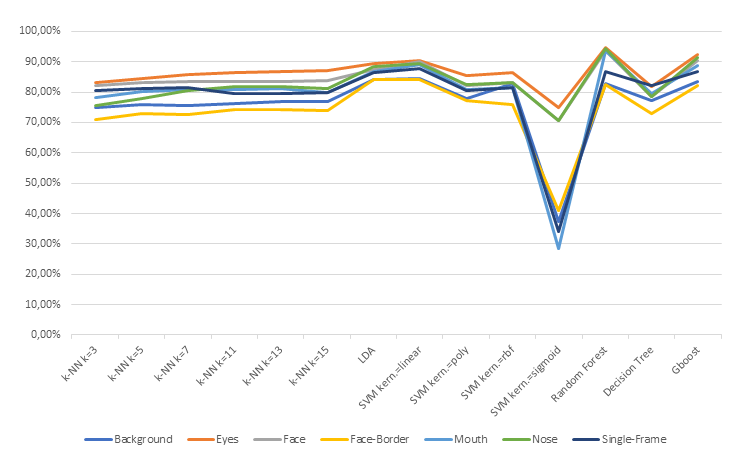}
    \caption{Values of AUC metric (\%) across classifiers and regions under analysis.}
    \label{fig:ROC}
\end{figure*}

Given a generic video $V$, the following feature vector was then created for each patch $p_i$ ($i={1,...,k}$, \textit{k = total number of patches extracted in video V}) of the area of interest $a \in A$: 

\begin{equation}
    p_i^{V_a} = \{\beta_1, \beta_2, \dots, \beta_{63}\}
\end{equation}

In order to obtain a unique descriptor $\overrightarrow{p^{V_a}}$ of $a$ for video $V$, the average of each $\beta$ component was computed, thus obtaining the following feature vector:

\begin{equation}
    p^{V_a} = \{\bar{\beta_1},\bar{\beta_2}, \dots, \bar{\beta_{63}}\}
\end{equation}

Different classifiers were trained on each $a$ area, giving feature $\overrightarrow{p^{V_a}}$ as input in order to identify the most discriminative region capable of solving the task in question. The classifiers used are:

\begin{itemize}
    \item k-NN with k = \{3, 5, 7, 11, 13, 15\};
    \item Linear Discriminative Analysis (LDA);
    \item  Support Vecotr Machine (SVM) with kernel = \{linear, poly, RBF, sigmoid\}
    \item Random Forest;
    \item Decision Tree; 
    \item GBoost.
\end{itemize}

Figure~\ref{fig:p2} summarizes the proposed approach.

The proposed method was implemented using the Python language, OpenCV and the Pytorch framework.

\section{Experimental Results and Comparison}
\label{sec:exp}

In the experiments carried out, all feature vectors $\overrightarrow{p^{V_a}}$ (described in the previous section) were collected and given as input to the various classifiers. As an evaluation metric, we chose AUC (Area Under the Curve). 

Figure \ref{fig:ROC} shows the results obtained for each classifier with respect to the various areas $A$ under analysis.
It can be seen from the Figure that the best results in the task in question were obtained using the Random Forest classifier. 
Specifically, the most discriminative areas in both datasets turn out to be the eyes with an AUC value equal to \textbf{94.55\%} and the nose with an AUC value of \textbf{94.28\%}.
The least discriminative regions turn out to be the facial border and the background. 
\begin{table}[t!]
\centering
\begin{tabular}{cc}
\hline
\textbf{Methods} & \textbf{AUC (\%)} \\ \hline
Visual Artifacts~\cite{matern2019exploiting} & 66,55             \\ \hline
Multi-task~\cite{nguyen2019multi}       & 66,3              \\ \hline
Face+Context~\cite{nirkin2021deepfake}     & 70,5              \\ \hline
LipForensics~\cite{haliassos2021lips}     & 89,75             \\ \hline
TD-3DCNN~\cite{zhang2021detecting}         & 80,52             \\ \hline
FTCN~\cite{zheng2021exploring}            & 93.3              \\ \hline
\textbf{Our}     & \textbf{94,55}    \\ \hline
\end{tabular}
\caption{Tab.2: AUC Score (\%) Comparisons with State-of-the-Art Approaches on FaceForensics++, and Celeb-DF(v2) Datasets.}
\label{tab:Approccio}
\end{table}

In general, we can therefore affirm that by exploiting the $\beta$ values extracted from the AC coefficients from the nose and eyes and analyzing only the type I frames, the proposed method is able to discriminate well and solve the task in question, despite not being a deep learning approach.

The best results obtained from the proposed approach were compared with state-of-the-art methods, many of which are based on deep learning algorithms. The AUC matric was calculated for each method. Table \ref{tab:Approccio} shows the obtained results. 
It can be seen from the obtained results that, the proposed method outperforms, albeit slightly, the state of the art. It should be highlighted, however, that the method we propose is fully analytical, explicable and fast in execution, compared to the various works in the literature. In addition, under these aspects, it is possible to use this method for real-time applications in order to counter the illicit use of the powerful deepfake technology.

\section{Conclusion}
\label{sec:conclusion}

In this study, we have proposed a novel deepfake detection algorithm that leverages the Discrete Cosine Transform (DCT) to extract discriminative features from video frames. By analyzing the $\beta$ components derived from the AC coefficients, we have identified informative frequencies that allow us to differentiate between real and deepfake videos effectively. Our approach prioritizes both accuracy and computational efficiency, aiming to enable real-time deepfake detection in practical scenarios.

To further enhance the efficacy of our algorithm, we have investigated discriminative patches within individual frames, focusing on regions like the eyes and mouth. Our analysis of these specific areas, particularly in I-frames, has revealed distinctive characteristics that differentiate real videos from deepfake ones. This patch-level analysis enhances the interpretability and reliability of our deepfake detection system.
Through extensive experiments on well-established deepfake datasets, such as Faceforensics++ and Celeb-DF (v2), we have demonstrated the effectiveness of our proposed approach in accurately classifying real and deepfake videos. 
By presenting our methodology, experimental findings, and analysis, this research significantly contributes to the field of deepfake detection. Our focus on speed, interpretability, and accuracy aims to advance the state-of-the-art in deepfake detection techniques. The insights gained from this study can aid in mitigating the potential harmful effects of deepfakes and contribute to the development of robust defense mechanisms against their malicious use.
While our proposed algorithm shows promising results, the constant evolution of deepfake algorithms calls for continued research and innovation in this domain. Future work should explore additional features and leverage advancements in machine learning and computer vision to further enhance the detection accuracy and efficiency of deepfake detection systems.

\section{Acknowledgments} 
This research is supported by Azione IV.4 - ``Dottorati e contratti di ricerca su  tematiche dell’innovazione" del nuovo Asse IV del PON Ricerca e Innovazione 2014-2020 “Istruzione e ricerca  per il recupero - REACT-EU”- CUP: E65F21002580005.




\small

\small
\bibliographystyle{fullname}
\bibliography{ref}


\begin{biography}

Luca Guarnera was born in Catania on October 26, 1992. Since January 1, 2022, he is a research fellow in Computer Science at the University of Catania. He graduated as Ph.D. in Computer Science (XXXIII cycle, PON number E37H18000330006) on October 14, 2021, discussing the thesis entitled “Discovering Fingerprints for Deepfake Detection and Multimedia-Enhanced Forensic Investigations” at the Department of Mathematics and Computer Science, University of Catania.  His main research interests are Computer Vision, Machine Learning, Multimedia Forensics and its related fields with a focus on the Deepfake phenomenon.

Salvatore Manganello was born in San Cataldo on October 14, 1998. He completed her bachelor's degree in Computer Science at the University of Catania. Her primary research areas focus on the forensic study of video deepfake detection.

Sebastiano Battiato is a full professor of Computer Science at the University of Catania. He received his degree in Computer Science (summa cum laude) in 1995 from the University of Catania and his Ph.D. in Computer Science and Applied Mathematics from the University of Naples in 1999. He has been Chairman of the Undergraduate Program in Computer Science (2012-2017), and Rector's delegate for Education: postgraduates and Phd (2013-2016). He is currently the Scientific Coordinator of the PhD Program in Computer Science (XXXIII-XXXVI cycles) and Deputy Rector for Strategic Planning and Information Systems at the University of Catania. His research interests include Computer Vision, Imaging technology and Multimedia Forensics.

\end{biography}

\end{document}